\documentclass{article}

\usepackage[final]{corl_2019} 

\usepackage{mathtools}
\usepackage{amsfonts}  

\usepackage{algorithmicx}
\usepackage[ruled]{algorithm}
\usepackage{algpseudocode}

\algnewcommand\algorithmicforeach{\textbf{for each}}
\algdef{S}[FOR]{ForEach}[1]{\algorithmicforeach\ #1\ \algorithmicdo}

\DeclarePairedDelimiterX{\infdivx}[2]{(}{)}{%
  #1\;\delimsize\|\;#2%
}

\newcommand{\kld}[2]{\ensuremath{D_{KL}\infdivx{#1}{#2}}}

\usepackage{expl3}
\ExplSyntaxOn
\newcommand\latinabbrev[1]{
  \peek_meaning:NTF . {
    #1\@}%
  { \peek_catcode:NTF a {
      #1.\@ }%
    {#1.\@}}}
\ExplSyntaxOff

\newcommand{\algmargin}{\the\ALG@thistlm}
\makeatother
\newlength{\whilewidth}
\algdef{SE}[parWHILE]{parWhile}{EndparWhile}[1]
  {\parbox[t]{\dimexpr\linewidth-\algmargin}{%
     \hangindent\whilewidth\strut\algorithmicwhile\ #1\ \algorithmicdo\strut}}{\algorithmicend\ \algorithmicwhile}%
\algnewcommand{\parState}[1]{\State%
  \parbox[t]{\dimexpr\linewidth-\algmargin}{\strut #1\strut}}
  
\let\OldStatex\Statex
\renewcommand{\Statex}[1][3]{%
  \setlength\@tempdima{\algorithmicindent}%
  \OldStatex\hskip\dimexpr#1\@tempdima\relax}

\usepackage{enumitem}
\usepackage{graphics}
\usepackage[pdftex]{graphicx}
\usepackage{subcaption}
\usepackage{caption}
\usepackage{wrapfig}

\title{Dynamics Learning with Cascaded Variational Inference for Multi-Step Manipulation}

\author{%
Kuan Fang$^{1}$, %
Yuke Zhu$^{1, 2}$, %
Animesh Garg$^{2,3}$, %
Silvio Savarese$^{1}$, %
Li Fei-Fei$^{1}$%
\\$^{1}$ Stanford University, %
$^{2}$ Nvidia, %
$^{3}$ University of Toronto \& Vector Institute
}

\begin{document}
\maketitle


\begin{abstract}
The fundamental challenge of planning for multi-step manipulation is to find effective and plausible action sequences that lead to the task goal. We present Cascaded Variational Inference (CAVIN) Planner, a model-based method that hierarchically generates plans by sampling from latent spaces. To facilitate planning over long time horizons, our method learns latent representations that decouple the prediction of high-level effects from the generation of low-level motions through cascaded variational inference. This enables us to model dynamics at two different levels of temporal resolutions for hierarchical planning. We evaluate our approach in three multi-step robotic manipulation tasks in cluttered tabletop environments given high-dimensional observations. Empirical results demonstrate that the proposed method outperforms state-of-the-art model-based methods by strategically interacting with multiple objects. See more details at \href{http://pair.stanford.edu/cavin}{\small \texttt{pair.stanford.edu/cavin}}


\end{abstract}


\section{Introduction}

Sequential problem solving is a hallmark of intelligence. Many animal species have demonstrated remarkable abilities to perform multi-step tasks~\cite{wimpenny2009cognitive,kohler2018mentality}. Nonetheless, the ability to solve multi-step manipulation tasks remains an open challenge for today's robotic research. The challenge involves high-level reasoning about what are the desired states to reach, as well as low-level reasoning about how to execute actions to arrive at these states. Therefore, an effective algorithm should not only make a high-level plan which describes desired effects during task execution, but also produce feasible actions under physical and semantic constraints of the environment. 

Conventional methods have formulated this as the task and motion planning problem~\cite{Kaelbling2010HierarchicalTA,Srivastava2014CombinedTA}. However, the applicability of these methods has been hindered by the uncertainty raised from visual perception in unstructured environments. To solve multiple tasks in such environments in a data-efficient manner, model-based methods, powered by deep neural networks, have been proposed to use data-driven dynamics models for planning~\cite{guo2014deep, Agrawal2016LearningTP, Finn2017DeepVF}. Given the trajectories predicted by the dynamics model, a plan can be generated through sampling algorithms, such as uniform sampling and cross entropy method~\cite{rubinstein2013cross}. These methods have shown successes in many control tasks given visual inputs, such as pushing objects in a cluttered tray~\cite{Finn2017DeepVF} and manipulating deformable objects~\cite{nair2017combining}. 


However, na\"{i}ve sampling approaches suffer from the curse of dimensionality when handling the large sampling space in manipulation domains. In realistic problems, we have to deal with continuous (and often high-dimensional) state and action spaces and long task horizons, whereas only a small fraction of actions is valid and effective and a small subset of state sequences could lead to high rewards. To boost sampling efficiency, recent work has proposed to use generative models~\cite{ichter2018learning, co2018self, chandak2019learning} to prioritize the sampling of more promising states and actions. These works do not exploit the hierarchical structure of multi-step tasks but rather making a flat plan on the low-level actions. As a result, the methods have mostly focused on short-horizon tasks.

\begin{figure*}[t!]
    \centering
    \includegraphics[width=\linewidth]{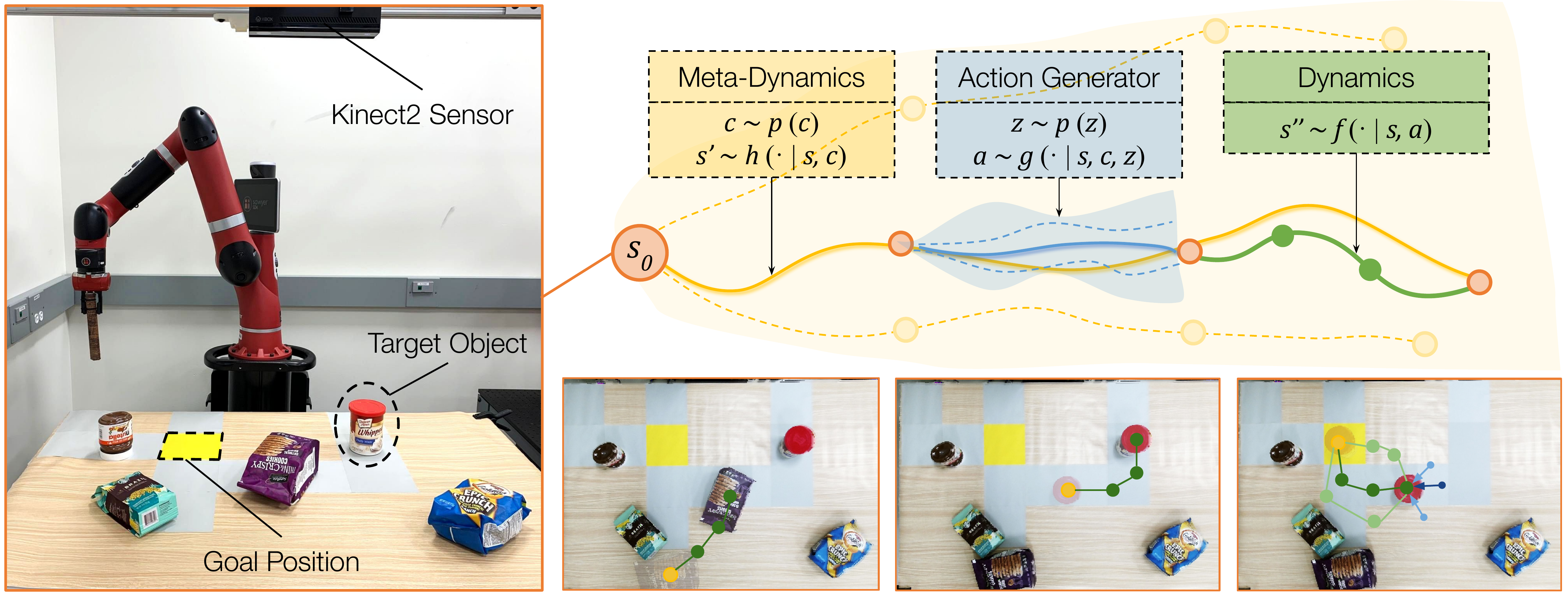}
    \caption{
    Hierarchical planning in latent spaces for multi-step manipulation tasks. The manipulation tasks shown in the figure requires the robot to move the target object to a goal position through specified regions (denoted by grey tiles). In presence of an obstacle, the planner needs to first clear the path and then move the target. We propose to use three tightly coupled modules: dynamics model, meta-dynamics model and action generator (see details in Sec.~\ref{sec:method}) to hierarchically generate plans for the task goal. Planning in learned latent spaces, our method first predicts subgoals (yellow) and then generates plausible actions (blue). The optimal plan is chosen by predicting resultant state trajectories (green) of the sampled actions. The selected plan is in darker colors. }
    \label{fig:main}
\end{figure*}

Our key insight for effective planning in multi-step manipulation tasks is to take advantage of the hierarchical structure of the action space, such that the generation of an action sequence can be factorized into a two-level process: 1) generating a high-level plan that describes the desired effects in terms of subgoals, and 2) generating a low-level plan of motions to produce the desired effects. To this end, we propose Cascaded Variational Inference (CAVIN) Planner to produce plans from learned latent representations of effects and motions. As illustrated in Fig.~\ref{fig:main}, the CAVIN Planner generates a high-level plan of desired effects predicted at a coarse temporal resolution and a low-level plan of motions at a fine-grained resolution. By decoupling effects and motions, the proposed method substantially reduces the search space of the optimal plan, enabling the method to solve longer-horizon manipulation tasks with high-dimensional visual states and continuous actions. To achieve this, we propose a cascaded generative process to jointly capture the distribution of actions and resultant states. We employ variational principles~\cite{kingma2013auto} in a cascaded manner to derive lower bound objectives for learning from task-independent self-supervision. 


Our contributions of this work are three-fold:
\begin{enumerate}[
    topsep=0pt,
    noitemsep,
    partopsep=0.5ex,
    parsep=0.5ex,
    leftmargin=*,
    itemindent=2.5ex
    ]
\item We introduce a model-based method using hierarchical planning in learned latent spaces. By performing predictive control (MPC) at two temporal resolutions, the proposed method substantially reduces the computational burden of planning with continuous action spaces over long horizons.

\item We propose a cascaded variational inference framework to learn latent representations for the proposed planner. The frameworks trains the model end-to-end with variational principles. Only task-agnostic self-supervised data is used during training. 

\item We showcase three multi-step robotic manipulation tasks in simulation and the real world. These tasks involve a robotic arm interacting with multiple objects in a cluttered environment for achieving predefined task goals. We compare the proposed approach with the state-of-the-art baselines. Empirical results demonstrate that our hierarchical modeling improves performance in all three tasks.  
\end{enumerate}

\setlength{\textfloatsep}{10pt}
\vspace{-10pt}
\section{Preliminaries}
\label{sec:problem_formulation}
\vspace{-5pt}
We consider robotic manipulation in an environment with unknown dynamics. Each manipulation task in the environment can be formulated as a Markov Decision Process (MDP) with high-dimensional state space $\mathcal{S}$ and continuous action space $\mathcal{A}$. We denote the dynamics as $p:\mathcal{S}\times\mathcal{A}\rightarrow\mathcal{S}$, where $p(s_{t+1}|s_t,a_t)$ is the transition probability from $s_t$ to $s_{t+1}$ by taking action $a_t$. In our tabletop environments, $s_t$ is the processed visual observations of the workspace from RGB-D cameras and $a_t$ is the control command of the robot (see Sec.~\ref{sec:experiments}). Our objective is to find a sequence of actions $a_{1:H}$ to maximize the cumulative reward $\mathbb{E}_{\pi}[\sum_{t=1}^{H} R(s_t, a_t, s_{t+1})]$ across planning horizon $H$, where $R:\mathcal{S} \times \mathcal{A} \times \mathcal{S}\rightarrow \mathbb{R}$ is the given immediate reward function. In this work, we consider the immediate reward function in the form of $R(s_{t+1})$ which can be directly evaluated from the next state. This enables planning in the state space while abstracting away actions.

We address the problem of finding the optimal action sequence $a^*_{1:H}$ with model-based methods. At the core of such an approach is a model predictive control (MPC) algorithm using a dynamics model of the environment~\cite{camacho2007mpc}. Given the known or learned dynamics model, we can predict resultant states of sampled actions by rolling out the actions with the dynamics model. The robot receives new observed state $s_t$ from the environment every $T$ steps and plans for the action sequence $a_{t:t+T-1}$. The planned actions are executed in a closed-loop manner for $T$ steps.
\vspace{-3pt}
\section{Method}
\label{sec:method}
\vspace{-3pt}
We present a hierarchical planning algorithm in learned latent spaces to prioritize promising plans. Our key insight is to factorize the generation of an action sequence into a two-level process: 1) generating a high-level plan that describes the desired \textit{effects} in terms of subgoals in the state space, and 2) generating a low-level plan of \textit{motions} that produces the desired effects. We propose Cascaded Variational Inference (CAVIN) to learn the model and the latent representations. 

\vspace{-3pt}
\subsection{Cascaded Generative Process of Dynamics}
\label{sec:generative_model}
\vspace{-3pt}
We propose to use a cascaded generative process to capture the distribution of actions and resultant states for efficient sampling-based planning. Instead of sampling in the original action space $\mathcal{A}$, we construct two latent spaces to facilitate the planning. We define $\mathcal{C}$ as the latent effect space, where an \textbf{effect code} $c\in\mathcal{C}$ describes the desired effect of the action, and $\mathcal{Z}$ as the latent motion space, where a \textbf{motion code} $z\in\mathcal{Z}$ defines the detailed motion of the action. 
Intuitively, each effect code represents a subgoal reachable from the current state in a $T$ steps, while each motion code represents a distinctive action sequence of $T$ steps that leads to the subgoal.
In our model, both $c$ and $z$ are continuous variables drew from priors as standard normal distribution $\mathcal{N}(0, I)$. 

Dynamics of the environment can be defined on two levels by introducing three tightly-coupled model components as shown in Fig.~\ref{fig:main}. On the low-level, we use the \textbf{dynamics model} $f$ to estimate the transition probability in the original state and action spaces. On the high-level, we use the \textbf{meta-dynamics model} $h$ to characterize the distribution of the subgoal state $s_{t+T}$ that is $T$ steps away from the current state $s_t$ with the effect code $c_t$. Compared to the low-level dynamics model, the meta-dynamics model operates at a coarser temporal resolution $T$ and abstracts away the detailed actions, which enables effective hierarchical planning in long-horizon tasks. The two levels of dynamics are related by the \textbf{action generator} $g$, which characterizes the distribution of the action sequences that will transit the current state $s_t$ into the chosen subgoal $s_{t+T}$ in $T$ steps. The action generator produces the action sequence $a_{t:t+T-1}$ using both of the effect code $c_t$ and the motion code $z_t$ given the current state $s_t$. Given the same $c_t$ and $s_t$, the meta-dynamics model and the action generator are supposed to produce consistent outputs. In other words, by taking action sequence $a_{t:t+T-1}$ generated by the action generator, the environment should approach the subgoal state $s_{t+T}$ predicted by the meta-dynamics model.  


Formally, the generative process of the two-level dynamics can be written as:
\begin{eqnarray}
    s_{t+1} & \sim & f(\cdot | s_t, a_t; \theta_f) \label{eqn:dynamics} \\
    s_{t+T} & \sim & h(\cdot | s_t, c_t; \theta_h) \label{eqn:meta_dynamics} \\
    a_{t:t+T-1} & \sim & g(\cdot | s_t, c_t, z_t; \theta_g) \label{eqn:action_generator}
\end{eqnarray}
where $f$, $h$ and $g$ are Gaussian likelihood functions parameterized by neural networks with parameters $\theta_f$, $\theta_h$ and $\theta_g$ respectively. We implement the modules using relation networks~\cite{Santoro2017ASN}. 

\vspace{-3pt}
\subsection{Hierarchical Planning in Latent Spaces}
\vspace{-3pt}
The hierarchical planning algorithm in the latent spaces $\mathcal{C}$ and $\mathcal{Z}$ is shown in Algorithm~\ref{algo:hierarchical_planning}. Every $T$ steps, the planning algorithm receives the current state $s_t$ and plans across the planning horizon $H$. We assume that $H$ is divisible by $T$. The action sequence $a_{t:t+T-1}$ is produced by the algorithm and executed in the environment in a closed-loop manner for $T$ steps. By choosing the sequences of effect and motion codes, the optimal action sequence $a^*_{t:t+T-1}$ is computed using the action generator described in Equation (\ref{eqn:action_generator}).



\begin{algorithm}[!t]
\caption{Hierarchical Planning with Cascaded Generative Processes}
\begin{algorithmic}[1]
\Require initial state $s_0$, planning horizon $H$, temporal abstraction $T$, number of plan samples $N$
\State $K = H / T$
\State $t = 0$
\While{episode not done}
    \State Receive the new $s_{t}$ from the environment.
    
    \State Sample $N$ sequences of effect codes of length $K$ as $\{ c_{1:K}^{i} \}_{i=1}^{N}$. 
    
    \State Predict subgoals $\{ s_{t+T}^{i}, s_{t+2T}^i, \ldots, s_{t+KT}^{i} \}_{i=1}^{N}$ by recursively using Equation (\ref{eqn:meta_dynamics}) with \par 
    \hskip\algorithmicindent $\{ c_{1:K}^{i} \}_{i=1}^{N}$ and interpolate the state trajectory.
    
    \State Choose $c_{1:K}^{*}$ and corresponding subgoals with the highest cumulative reward $\sum_{t=1}^T R(s_{t})$.
    
    \State Sample $N$ sequences of motion codes of length $K$ as $\{ z_{1:K}^{j} \}_{j=1}^{N}$.
    
    \State Predict sequences of actions $\{ a_{t}^{j}, ..., a_{t+H-1}^{j} \}_{j=1}^{N}$ and states $\{ s_{t+1}^{j}, ..., s_{t+H}^{j} \}_{j=1}^{N}$ with \par
    \hskip\algorithmicindent $c_{1:K}^{*}$ and $\{ z_{1:K}^{j} \}_{j=1}^{N}$ by recursively using Equation (\ref{eqn:action_generator}) and (\ref{eqn:dynamics}).
    
    \State Choose $z_{1:K}^*$ and corresponding $a_{t:t+H-1}^*$ which lead to states closest to the subgoals.
    \State Execute $a_{t:t+T-1}^*$ for $T$ steps.
    \State Update $t = t + T$.
\EndWhile

\end{algorithmic}
\label{algo:hierarchical_planning}
\end{algorithm}
\setlength{\textfloatsep}{10pt}

\textbf{Effect-Level Plan}. On the high-level, we plan for subgoals towards reaching the final task goal by choosing the optimal sequence of effect codes $c^*_{1:K}$, where $K = T / H$ is the number of subgoals. $N$ sequences of effect codes are sampled from the prior probability $p(c)$, which we denote as $\{ c_{1:K}^{i} \}_{i=1}^{N}$. Using Equation~\eqref{eqn:meta_dynamics}, the trajectory of subgoals corresponds to each $c_{1:K}^{i}$ are recurrently predicted as $s_{t+T}^{i}, s_{t+2T}^i, \ldots, s_{t+KT}^{i}$. During this process, episodes which fail early will be replaced by duplicating episodes that are still active. Between every two adjacent subgoals, we linearly interpolate the states of each time step and evaluate the immediate reward $R(s_t)$. We use the cumulative rewards to rank the predicted sequence of states. The highest-ranked sequence $c_{1:K}^{*}$ and the corresponding subgoals are selected to serve as intermediate goals in the low-level planning. 

\textbf{Motion-Level Plan}. On the low-level, we generate the sequence of actions in the context of the desired effects indicated by $c_t$, by choosing the optimal sequence of motion codes $z^*_{1:K}$. $N$ sequences of motion codes are sampled from the prior probability $p(z)$ as $\{ z_{1:K}^{j} \}_{j=1}^{N}$. Each sequence is paired with the selected $c_{1:K}^{*}$ to produce the action sequence. Each pair of effect and motion codes is projected to a segment of action sequence of $T$ steps using the action generator $g$ as in Equation~\eqref{eqn:action_generator}. Then the resultant state trajectories are predicted by the dynamics model $f$ using Equation~\eqref{eqn:dynamics}. We choose $z_{1:K}^{*}$ which leads to states that are closest to the chosen subgoals in the high-level planning.

The optimal action sequence $a_{1:H}^{*}$ is produced as the one that corresponds to the selected $c_{1:K}^{*}$ and  $z_{1:K}^{*}$. The first $T$ steps of planned actions are provided to the robot for execution. Re-planning occurs every $T$ steps when the new state is received. 

\vspace{-3pt}
\subsection{Learning with Cascaded Variational Inference}
\label{sec:training}
\vspace{-3pt}
The dynamics model $f$, meta-dynamics model $h$ and action generator $g$ are learned with respect to parameters $\theta_h$, $\theta_f$ and $\theta_g$ by fitting the transition data $\mathcal{D} = \{(s^{i}_t, a^{i}_t, s^{i}_{t+1})\}_{i = 1}^{M}$ observed from the environment. $\mathcal{D}$ can be collected either by self-supervision from the robot's own experiences or human demonstrations. In this paper, we use a heuristic policy that randomly samples actions that are likely to have plausible effects to the environment. The learning algorithm aims to maximize the marginal likelihoods $p(s_{t+1}| s_t, a_t; \theta_f)$, $p(s_{t+T}| s_t; \theta_h)$, and $p(a_{t:t+T-1}|s_{t}, c_{t}; \theta_g)$ on the dataset $\mathcal{D}$ under the entire generative process as described in Sec.\ref{sec:generative_model}. 

For the low-level dynamics, the likelihood is directly maximized by observed tuples of $s_t$, $a_t$ and $s_{t+1}$ drew from $\mathcal{D}$. We define  $f(s_{t+1}| s_t, a_t; \theta_f)$ as a function which predicts the mean of $s_{t+1}$ with a fixed covariance matrix, so maximizing the likelihood is equivalent to minimizing a reconstruction loss between the observed $s_{t+1}$ and the prediction.

The meta-dynamics model and the action generator are trained with transition sequences of consecutive $T$ steps from $\mathcal{D}$. As the effect code $c$ and the motion code $z$ are latent variables, the likelihoods $p(s_{t+T}| s_t; \theta_h)$ and $p(a_{t:t+T-1}|s_{t}, c_{t}; \theta_g)$ cannot be directly maximized. Instead, we follow the variational principle~\cite{kingma2013auto} to derive a lower bound objective on these two marginal likelihoods by constructing inference models of $c$ and $z$. For succinctness, we drop the subscripts from the symbols in the following equations using $s'$ to denote the next subgoal state $s_{t+T}$, $s''$ to denote the next state $s_{t+1}$, and $a$ to denote the action sequence $a_{t:t+T-1}$.

\noindent
\textbf{Lower Bound Objective.}
We construct inference models which consist of approximate posterior distributions $q_h(c | s, s'; \phi_h)$ and $q_g(z | s, a, c; \phi_g)$ both as inference neural networks with trainable parameters denoted as $\phi_h$ and $\phi_g$. Thus the evidence lower bound objective (ELBO)~\cite{kingma2013auto} for marginal likelihoods can be derived as below.



For the meta-dynamics $h$, the variational bound $\mathcal{J}_h$ on the marginal likelihood for a single transition is defined as a standard conditional variational autoencoder (CVAE)~\cite{kingma2013auto} conditioned on $s$:
\begin{equation}
    \log p(s' | s; \theta_h) \geq \mathbb{E}_{q_h(c | s, s'; \phi_h)}[\log h(s' | s, c; \theta_h)] - \kld{q_h(c | s, s'; \phi_h)}{p(c)} = - \mathcal{J}_h
\end{equation}

For the action generator $g$, directly maximizing $p(a|s, c; \theta_g)$ conditioned on the unobserved $c$ is intractable. Instead, we maximize $p(a | s, s'; \theta_g)$ by marginalizing over the inferred $c$ given observed transitions from $s$ to $s'$:
\begin{eqnarray}
    p(a | s, s'; \theta_g)
     =  \mathbb{E}_{q_h(c | s, s'; \phi_h)} [p(a | s, s', c; \theta_g)]
     =  \mathbb{E}_{q_h(c | s, s'; \phi_h)} [p(a | s, c; \theta_g)]
\end{eqnarray}
Assume $c$ is given, the variational bound $\mathcal{J}_{g | c}$ of the marginal likelihood $p(a | s, c; \theta_g)$ is:
\begin{equation}
    \log p(a | s, c; \theta_g)
    \geq \mathbb{E}_{q_g(z | s, a, c; \phi_g)}[\log g(a | s, c, z; \theta_g)] - \kld{q_g(z | s, a, c; \phi_g)}{p(z)}
    = - \mathcal{J}_{g | c}
\end{equation}
Using Jensen's inequality, the variational lower bound $\mathcal{J}_g$ of marginal likelihood $p(a | s, s')$ can be derived as:
\begin{equation}
    \log p(a | s, s'; \theta_g)
    \geq \mathbb{E}_{q_h(c | s, s'; \phi_h)} [\log p(a | s, c; \theta_g)]
    \geq \mathbb{E}_{q_h(c | s, s'; \phi_h)} [- \mathcal{J}_{g | c}]
    = - \mathcal{J}_g
\end{equation}

The maximum likelihood estimation problem now becomes minimizing the objective function $\mathcal{J}_h + \mathcal{J}_g$ on $\mathcal{D}$ with respect to parameters $\theta_h$, $\theta_g$, $\phi_h$ and $\phi_g$ end-to-end.

\vspace{-3pt}
\section{Experiments}
\label{sec:experiments}
\vspace{-3pt}
We design our experiments to investigate the following three questions: 
1) How well does our method perform on different multi-step manipulation tasks? 
2) How important does it perform in various scene complexities? 
3) What kind of robot behaviors does our method produce?

\vspace{-3pt}
\subsection{Experimental Setup}
\label{sec:setup}
\vspace{-3pt}
\textbf{Environment.} We construct a simulated platform to evaluate multi-step manipulation tasks using a real-time physics simulator~\cite{coumans2017bullet}. As shown in Fig.~\ref{fig:main}, the workspace setup includes a 7-DoF Sawyer robot arm, a table surface, and a depth sensor (Kinect2) installed overhead. Up to 5 objects are randomly drawn from a subset of the YCB Dataset~\cite{alli2015TheYO} and placed on the table. The Sawyer robot holds a short stick as the tool to interact with the objects to complete a specified task goal. 

The observation consists of segmented point cloud represented by $m \times n \times 3$ Cartesian coordinates in the 3D space, where $m$ is the number of movable objects and $n = 256$ is the fixed number of points we sample on each object. The state is composed of the $m \times 3$ object centers and $m \times 64$ geometric features processed from the segmented point cloud, where the geometric features are extracted using a pretrained PointNet~\cite{qi2016pointnet}. The PointNet has three layers with dimensions of 16, 32, 64 and followed by a 64-dimensional FC layer. 

The robot performs planar pushing actions by position control. Each push is a straight line motion with maximum moving distance of 0.1 meters along x- and y-axes. The action of each step is defined as a tuple of coordinates which represents the initial and delta-positions of the robot end-effector. The planning horizon is $H = 30$ steps and each episode terminates after $60$ steps. Every $T = 3$ steps, the robot arm moves out of the camera view to take an unoccluded image for replanning.

\begin{figure*}[t!]
    \centering
    \includegraphics[width=\linewidth]{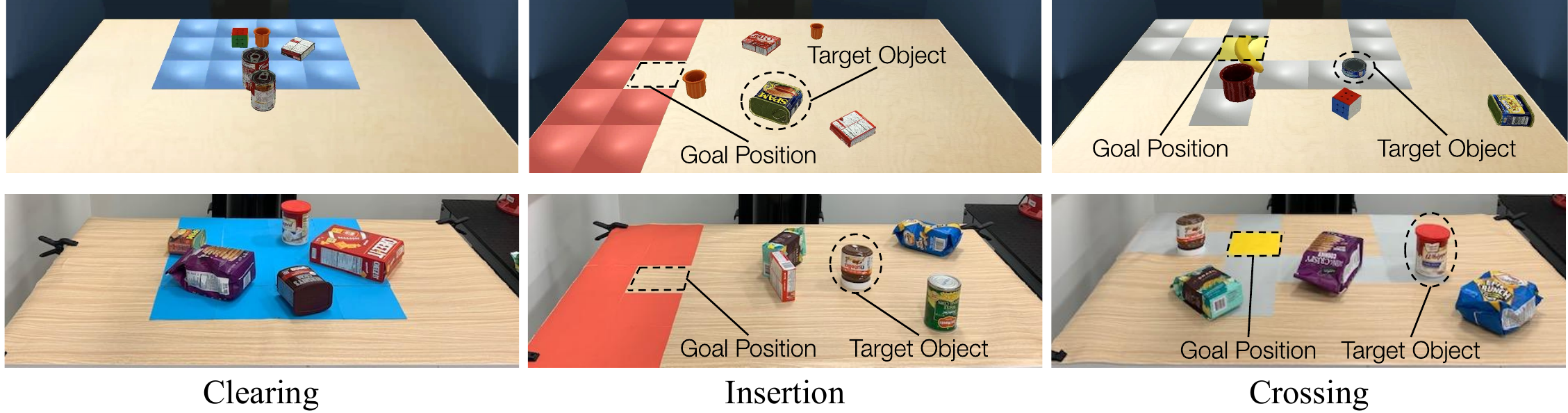}
    \vspace{-15pt}
    \caption{The environments of the three tabletop multi-step manipulation tasks in simulation and the real world. Objects and their initial placements are randomized in each episode.}
    \label{fig:tasks}
\end{figure*}

\textbf{Tasks.}
We design three multi-step manipulation tasks \textit{Clearing}, \textit{Insertion}, and \textit{Crossing}. A reward function is given to the robot for achieving the task goal. All tasks share the same MDP formulation in Sec.~\ref{sec:problem_formulation}, while the arrangement of the scene and the reward functions are constructed differently in each task. None of the tasks are seen by the model during training time. The three tasks are illustrated in Fig.~\ref{fig:tasks} with additional details in the Appendix.


In each task, dense and sparse reward functions are defined respectively. The dense reward function moderates the task complexity by providing intermediate reward signals, such as the Euclidean distance to the goal. While the sparse reward function only returns a positive value at the end of task completion, which poses a more intense challenge to plan ahead in a large search space requiring strategic interaction with different objects in diverse yet meaningful manner. 


\textbf{Baselines.} We compare our method with a set of baselines, all of which use learned dynamics model coupled with model-based planning by sampling actions from the original action space or a latent space. These include \textbf{MPC}~\cite{guo2014deep, Agrawal2016LearningTP, Finn2017DeepVF} which samples from original state and action spaces, \textbf{CVAE-MPC}~\cite{ichter2018learning} which learns a flat generative model to sample actions, and \textbf{SeCTAR}~\cite{co2018self} which learns a latent space model for states and actions for planning. More details are in the Appendix. 


\vspace{-10pt}
\subsection{Quantitative Comparisons}
\vspace{-3pt}
Our method and baselines are compared across the three different multi-step manipulation tasks. We evaluate each method in each experiment below with $1,000$ episodes given the same random seed. Each method draws $1,024$ samples in every planning step. We analyze the task success rate of each method in the three tasks by providing dense or sparse rewards and varying the scene complexity. 

\begin{figure*}[t!]
    \centering
    \begin{subfigure}[t]{0.32\textwidth}
        \centering
        \includegraphics[width=\linewidth]{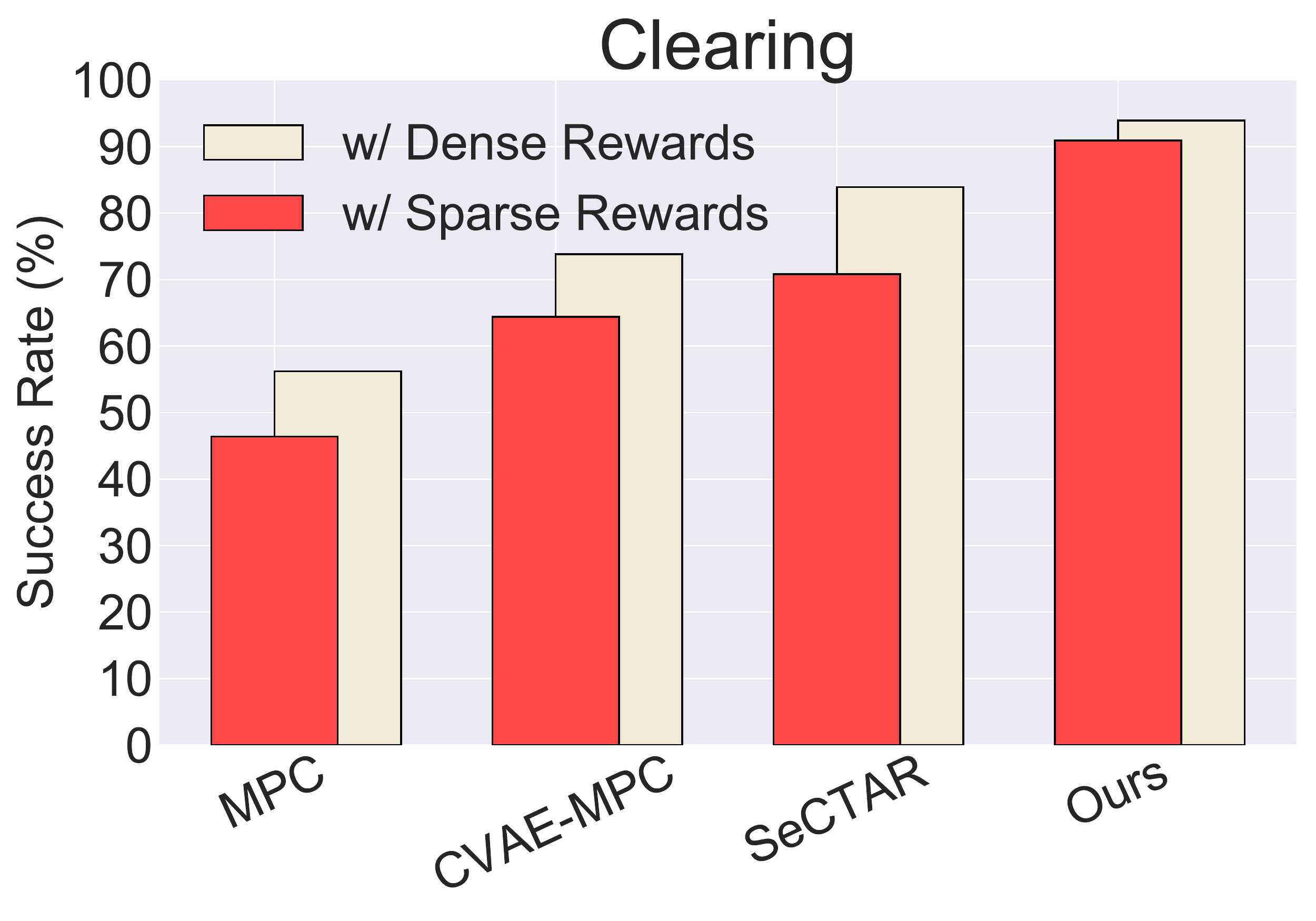}
    \end{subfigure}
    \begin{subfigure}[t]{0.32\textwidth}
        \centering
        \includegraphics[width=\linewidth]{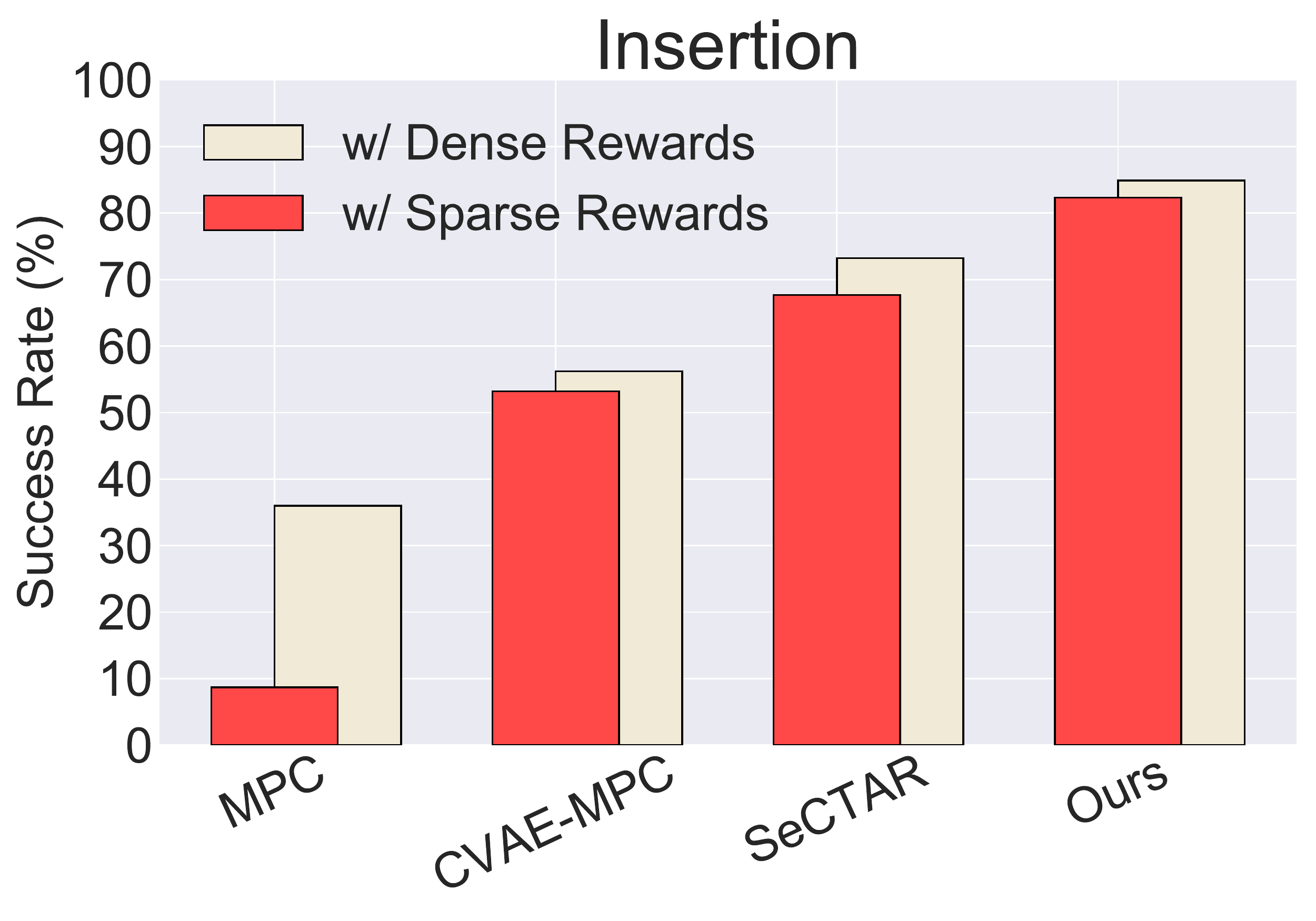}
    \end{subfigure}
    \begin{subfigure}[t]{0.32\textwidth}
        \centering
        \includegraphics[width=\linewidth]{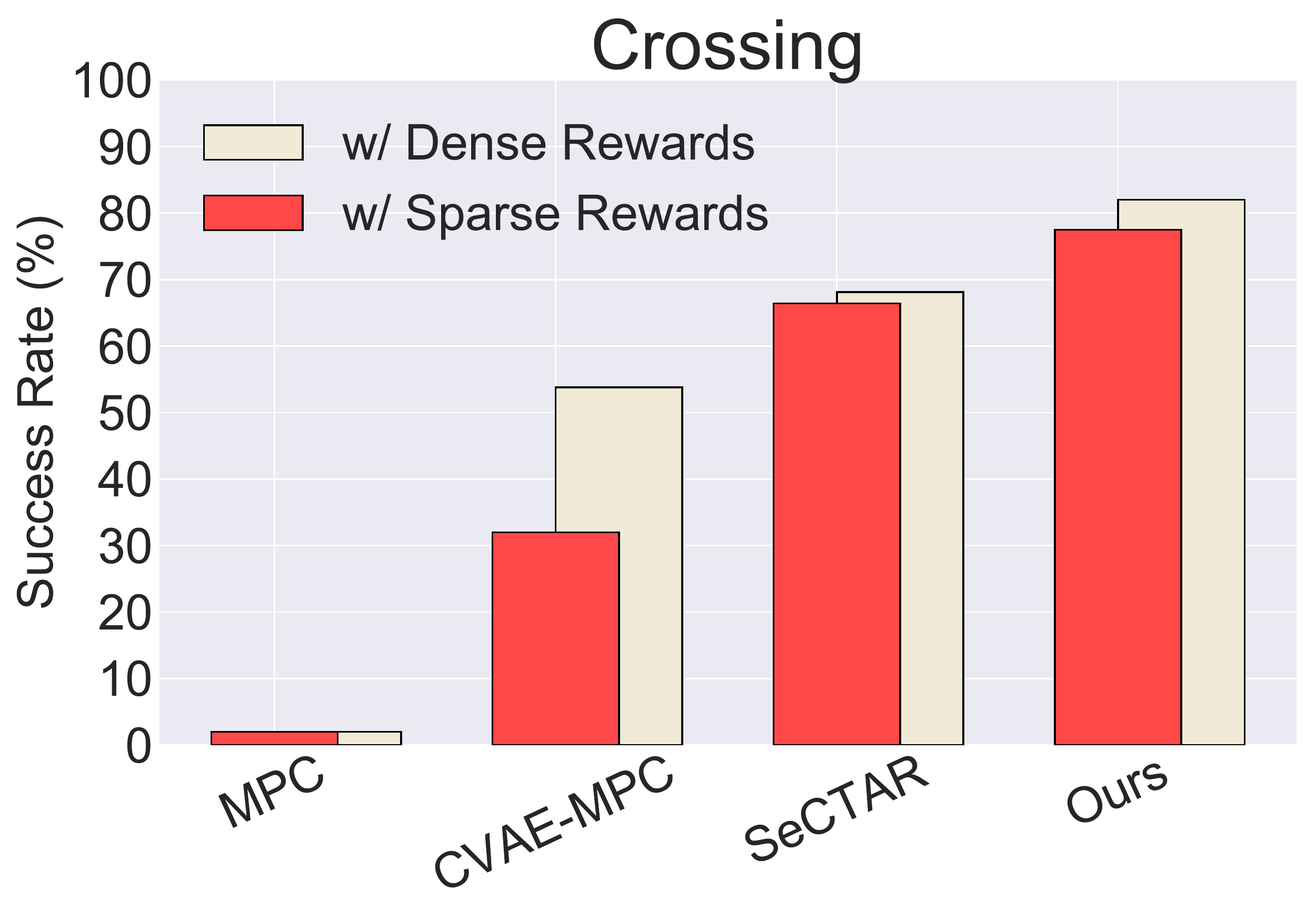}
    \end{subfigure}
    \begin{subfigure}[t]{0.32\textwidth}
        \centering
        \includegraphics[width=\linewidth]{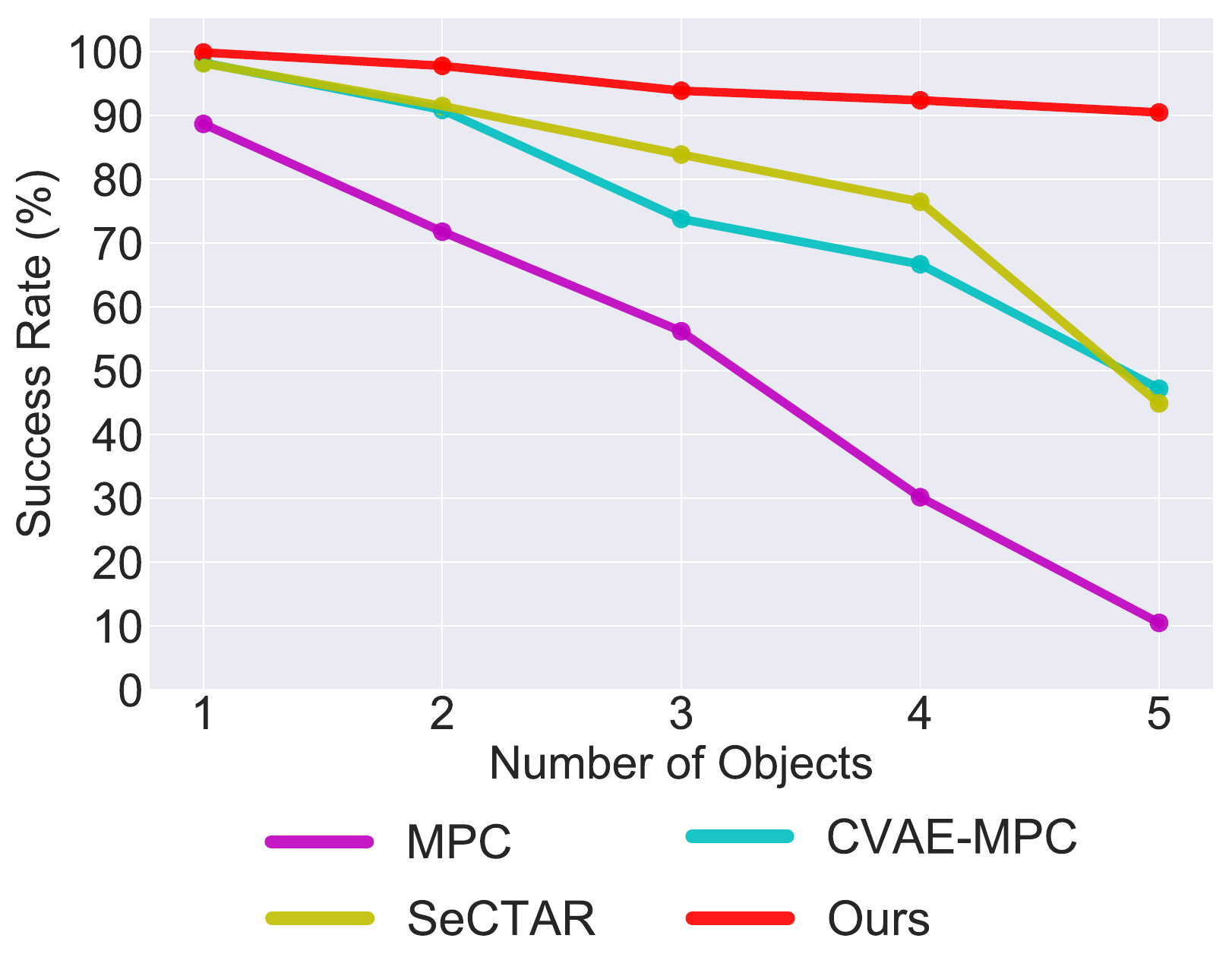}
    \end{subfigure}
    \begin{subfigure}[t]{0.32\textwidth}
        \centering
        \includegraphics[width=\linewidth]{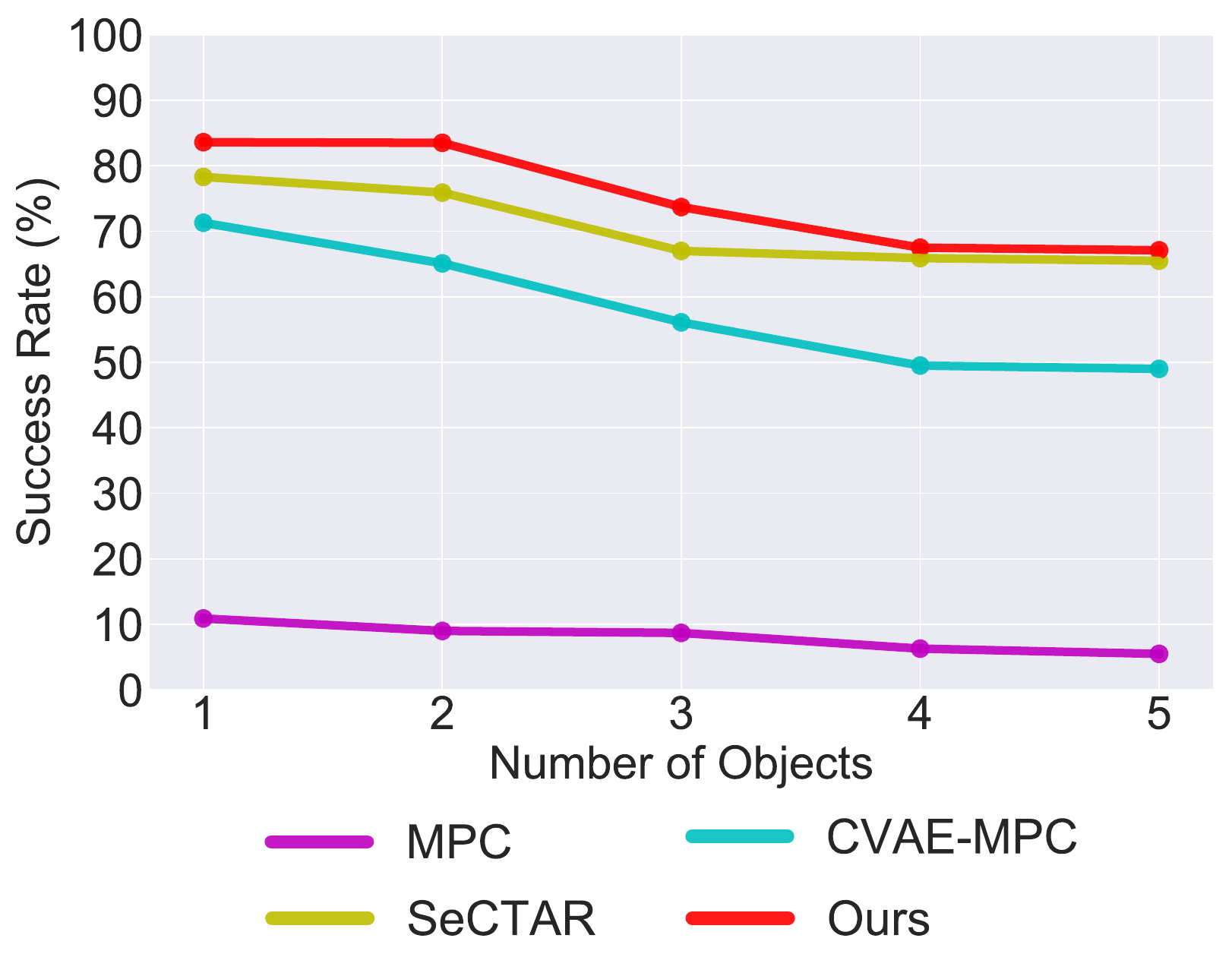}
    \end{subfigure}
    \begin{subfigure}[t]{0.32\textwidth}
        \centering
        \includegraphics[width=\linewidth]{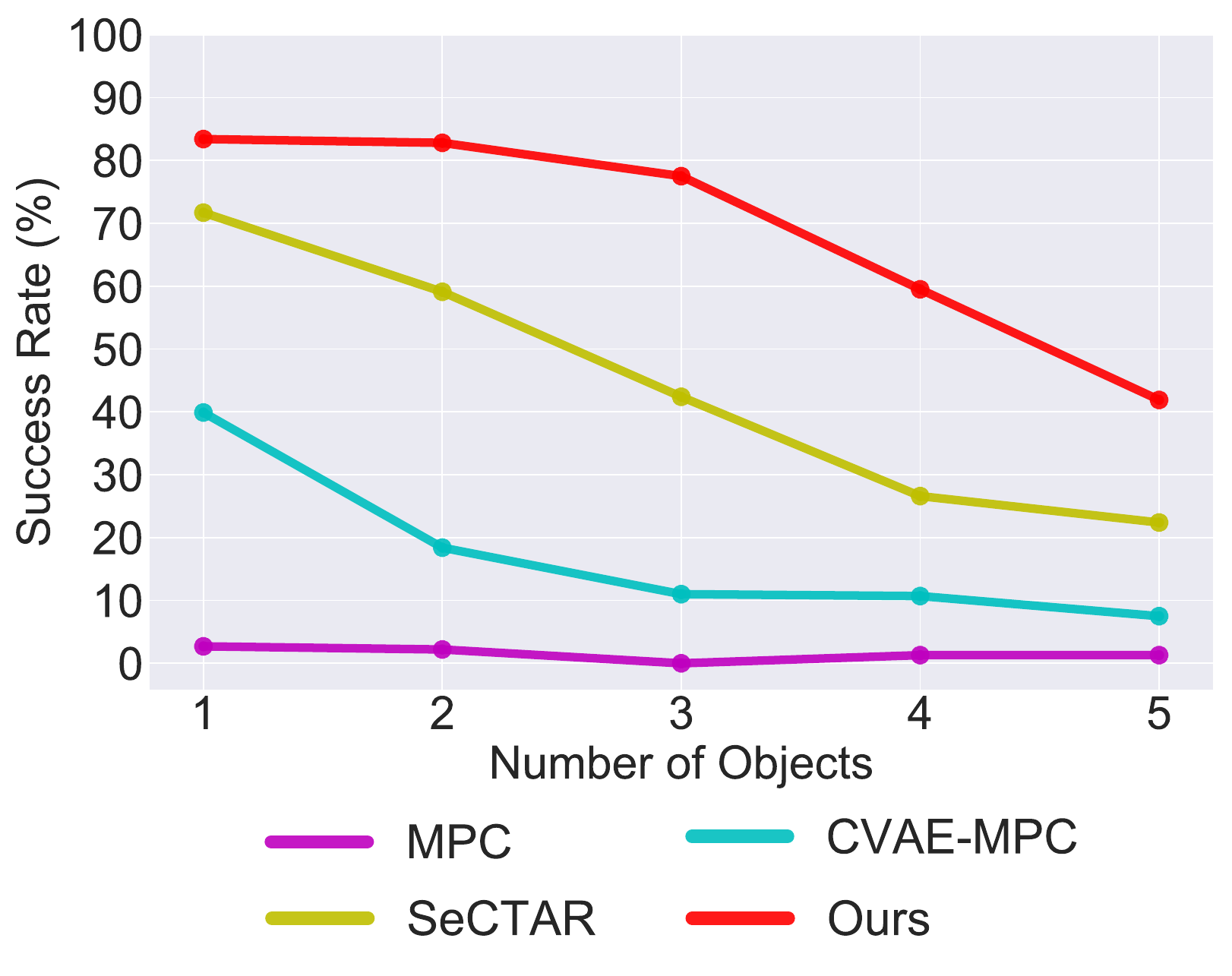}
    \end{subfigure}
    \caption{\textbf{Top Row:} Task performance with 3 objects given Dense and Sparse Rewards. In both cases, our method consistently outperforms all baseline methods. Compared to planning methods without hierarchies, our method is more robust under sparse rewards.
    \textbf{Bottom Row:} Task performance against number of objects. We evaluate the models' scalability and robustness with growing number of objects. We demonstrate that our method suffers the least from the increased complexity due to the cascaded generation of actions.}
    \label{fig:performance_vs_object}
\end{figure*}

\textbf{Task performance given dense and sparse rewards.} We compare all methods with dense and sparse rewards as shown in Fig.~\ref{fig:performance_vs_object}(top row). The number is evaluated with 3 objects initialized in each environment: one target object and two obstacles in \texttt{Insertion} and \texttt{Crossing}. Across all tasks, our method outperforms all other model-based baselines under both dense and sparse rewards. Given the dense reward functions, it has considerable margins compared to the second best methods (10.0\% in \texttt{Clearing}, 11.7\% in \texttt{Insertion}, 13.9\% in \texttt{Crossing}). Especially for the latter two tasks which requires longer-term planning, the performance gap is telling. The planning becomes harder under sparse rewards, where a na\"{i}ve or greedy algorithm cannot easily find the good strategy. In this case, performances of most methods drops under sparse rewards. However, our method suffers less than 5\% drop across all 3 tasks. The margins over the second best are 20.1\% in \texttt{Clearing}, 14.6\% in \texttt{Insertion}, 11.1\% in \texttt{Crossing}. This result demonstrates a strong advantage of the multi-scale hierarchical-dynamics model of our method for long-term planning under sparse rewards. 

Compared to the baselines, our method efficiently rules out trajectories of unsuccessful effects in the high-level planning process and effectively finds the reachable sequence of states that will lead to the task goal. In the low-level planning, our method focuses on sampling actions that lead to such effects. We also observe that different baseline fail for different reasons. Intuitively, in a large search space, uniformly sampled actions are largely ineffective or infeasible. Therefore, only a small fraction of samples in MPC actually lead to plausible solutions, while others are fruitless. CVAE-MPC uses the action generator in a flat planning framework. It produces task performance comparable to our method with dense rewards which provide intermediate guidance for correction. But under the sparse rewards, its performance is significantly undermined due to increasing difficulty of finding a long-horizon plan. SeCTAR effectively eliminates the undesired sequence of states using the meta-dynamics model similar to our method. However, using an entangled latent space for generating both states and actions leads to poor quality of generated action samples. While out method avoids such problem by introducing an additional latent variable as the motion code. Empirically, we also found the training of SeCTAR requires a careful balance between the reconstruction losses of state and action in order to yield reasonable results, which is a challenge in itself.

\textbf{Task performance against number of objects.} Given sparse rewards, we vary the number of objects in the workspace to evaluate the model performance under different task complexity as shown in Fig.~\ref{fig:performance_vs_object}. 
More objects lead to a exponentially growing search space of feasible plans, due to the combinatorial nature of subtask dependencies. We see that increasing the number of objects leads to performance drop in all methods. However, our method shows the highest robustness and maintains the best performance. Especially in \texttt{Clearing}, the performance of our method only drops by 9.4\% when increasing the number of objects from 1 to 5. While the performances of baseline methods drops 51.1\%, 53.3\% and 70.2\% respectively. 

\textbf{Real-world experiments.} We also evaluate our method in the real-world environments for the three tasks. The environment setup and the reward functions are identical to the simulated experiments. Since the observations of these tasks are based on point clouds which has relatively little reality gap, we can directly apply the model trained in the simulation in the real world without explicit adaptation. In each episode, we randomly initialize the environment with real-world objects including packaged foods, metal cans, boxes and containers. Among 15 evaluated episodes, out method achieves success rates of 93.3\% in \texttt{Clearing}, 73.3\% in \texttt{Insertion}, and 80.0\% \texttt{Crossing}, which is comparable to our simulated experiments. 



\subsection{Qualitative Results}

\begin{figure*}[t!]
    \centering
    \includegraphics[width=\linewidth]{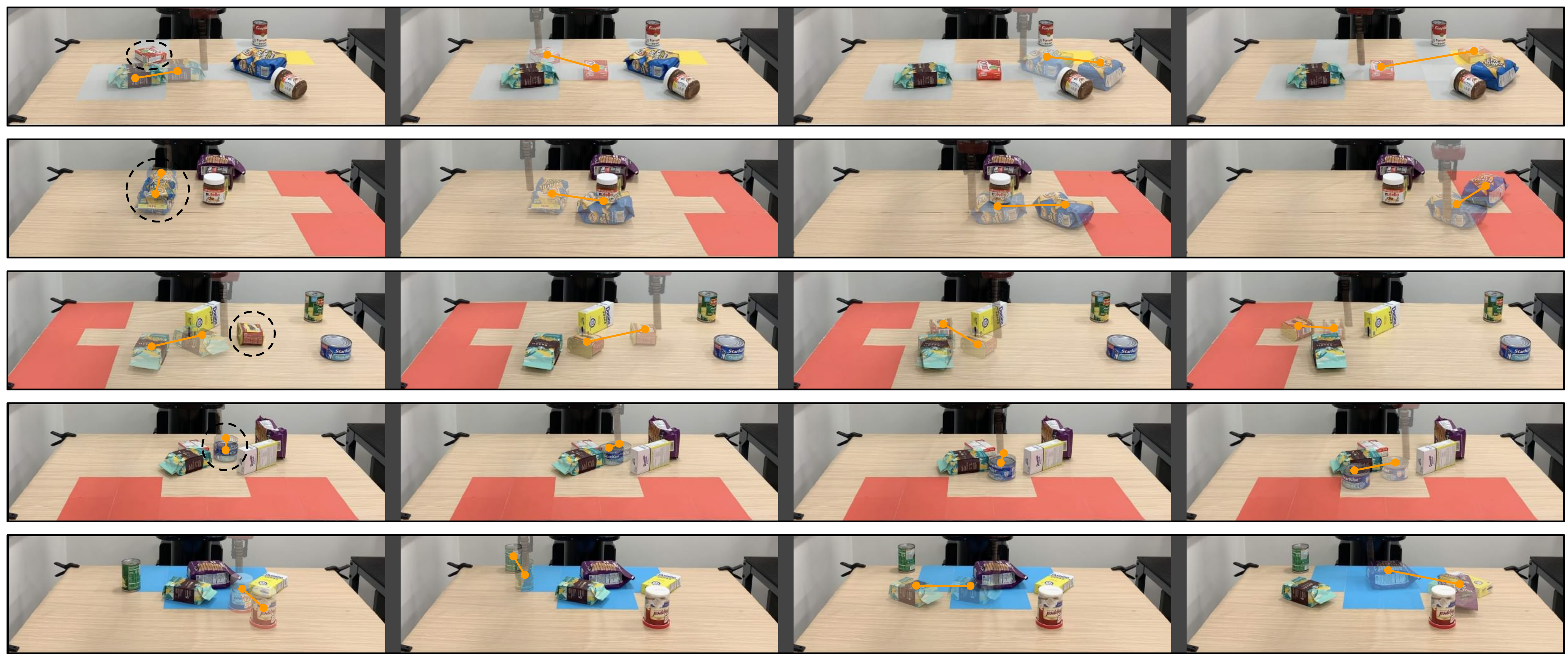}
    \caption{Visualization of the task execution in the real world using our trained model. Each row shows a different episode in temporal order from left to right. The subgoal is overlaid on the current state at each step and connected by an orange line. The target objects are marked by dashed circles.}
    \label{fig:qualitative}
\end{figure*}

In Fig.~\ref{fig:qualitative}, we visualize the planned trajectories predicted by our method. Example episodes of the three tasks are shown in each row from left to right. We observe that the robot adopts diverse strategies planned by our method. In presence of obstacles, the robot moves each obstacle aside along the path of the target object. Given a pile of obstacles between the target and the goal, the robot pushes the target around. When the target is surrounded by several obstacles, the robot opens a path for it towards the goal. When the target object is small and there is a small gap on its way towards the goal, the robot squeezes the target through the gap. To clear multiple objects in a region, the robot moves the objects away one by one. Many of these behaviors require strategic interactions with multiple objects in a specific order.

\vspace{-3pt}
\section{Related Work}
\label{sec:related_works}
\vspace{-3pt}

\textbf{Hierarchical Planning.} Learning diverse skills for hierarchical policies and controller is a common practice in control and robotics. One way to learn a set of different skills is to pre-speficy a set of modes and learn a policy for each mode. Each policy has a different parameterization and trained for a specific purpose. \cite{andreas2017modular} composes subpolicies of different action modes using policy sketches provided beforehand and trains the composed policies as modular neural networks~\cite{andreas2016neural}.
In robot manipulation, several methods define separate grasping and manipulation policies for pushing/tool use \cite{zeng2018learning, fang2018tog, danielczuk2019mechanical}. In contrast, our approach does not assume the action modes are specified beforehand. And our training procedure uses self-supervised interactions to learn a hierarchy of latent effects and motions, instead of relying on pre-specified low-level policy primitives.

\textbf{Skill Learning.} Another line of work aims to discover action modes from the data in an unsupervised manner. \cite{Thrun1994FindingSI} defines skills as subpolicies that apply to different subsets of states, which can be learned by minimizing the Q-learning objective. Several multi-modal imitation learning algorithms propose to learn diverse behaviors from demonstrations \cite{li2017infogail, hausman2017multi, tamar2018imitation}. These works assume the demonstration data is generated by multiple expert policies, which can be imitated separately in an adversarial learning framework. Our approach shares a similar goal as we aim to discover the action modes from data. While these works factorize action modes according to trajectories or subsets of states, our approach learns the structure of the action spaces by \textit{how they affect} the environment. 

\textbf{Generative Models for Planning.} In addition to the baselines discussed in Sec.~\ref{sec:experiments}, Visual MPC~\cite{Finn2017DeepVF}, CVAE-SBMP~\cite{ichter2018learning}, and SeCTAR~\cite{co2018self}, several other methods can be used for dynamics model learning. 
\cite{ichter2018learning} uses conditional variational autoencoder (CVAE) \cite{kingma2013auto, sohn2015learning} to learn a generative model to draw collision-free samples from the action space. This idea is further extended for collision-free motion planning \cite{ichter2019robot}. Co-Reyes et al.~\cite{co2018self} use a policy decoder to generate sequence of actions and a state decoder to predict the dynamics given the same latent feature as the input. The two decoders are jointly trained to encourage the predicted future states are consistent with the resultant states caused by the generated actions. Both decoders receive the same latent feature as input. The main notable difference is that most of these methods represent the probability distribution of a single action mode, where the data is often deliberate for the task. While, the hierarchical dynamics model in the CAVIN Planner decouples the model learning into latent code for effects and motion codes, each of which can guide the action sampling. And finally the consistency action sampling is ensured through dynamics prediction over a self-supervised dataset.


\section{Conclusion}

In this work, we propose a hierarchical planning framework using learned latent spaces for multi-step manipulation tasks. Our model improves the efficiency of planning in complex tasks with high-dimensional continuous state and action spaces by exploiting the hierarchical abstraction of action spaces. Our model hierarchically performs model-based planning at two different temporal resolutions: the high-level prediction of effects and the low-level generation of motions. And the model is learned from task-agnostic self-supervision using cascaded variational inference. We evaluate our method in three long-horizon robotic manipulation tasks with high-dimensional observations. The experiment results indicate that our method outperforms strong model-based baselines by large margins, as a result of its efficient generation of plausible action sequences.

\textbf{Acknowledgement:} We acknowledge the support of Toyota (1186781-31-UDARO). We thank Ajay Mandlekar and Matthew Ricks for the infrastructures in real-world experiments, Kevin Chen, Eric Yi and Shengjia Zhao for helpful discussions, and anonymous reviewers for constructive comments.

\clearpage
\renewcommand*{\bibfont}{\footnotesize}
\renewcommand{\baselinestretch}{.95}

\newpage
\appendix
\section{Implementation Details}

\subsection{Network Architectures}
The architectures of dynamics model, meta-dynamics model and action generator in the CAVIN planner are shown in Fig.~\ref{fig:cavin_modules}. The design of these network architectures are implemented with relation networks~\cite{Santoro2017ASN} to aggregate information across objects. In each network, the feature of the current state $s_t$ is first computed as three $N \times 64$ arrays: spatial features, geometric features, and relation features. Spatial features and geometric features are computed from the center and the point cloud of each object using fully-connected (FC) layers. The relation features are computed by relation networks. The action $a_t$ or the effect code $c_t$ is processed by another FC layer and then concatenated with the state features. In dynamics and meta-dynamics models, the change of the state $\Delta s_t$ is computed from the concatenated features and added to the current state for each object. In the action generator, we need to aggregate $s_t$, $c_t$ and $z_t$ to compute the action sequence. After the features of the $s_t$ and $c_t$ are concatenated and processed, we pool the feature across all objects. We transform the distribution of $z_t$ using the pooled feature. Specifically, the mean $\mu$ and the variance $\Sigma$ are computed from the pooled feature to compute the transformed latent code $\mu + \Sigma \cdot z_t$. And the action sequence is computed from the transformed latent code.

The architectures of effect inference network and motion inference network are shown in Fig.~\ref{fig:inference_networks}. In the effect inference network, we first compute a per-object feature from the concatenation of $s_t$ and $s_{t+1} - s_t$ using FC layers. Then we pool the feature across objects. The mean $\mu_{c, t}$ and the covariance $\Sigma_{c, t}$ of the effect code $c_t$ is computed as the output. The motion inference network in the CAVIN Planner uses a similar architecture with the action generator to aggregate the information of $s_t$ and $c_t$. Then the mean $\mu_{z, t}$ and the covariance $\Sigma_{z, t}$ of the motion code $z_t$ are computed by the concatenation of the pooled feature and the action feature.
 
The latent codes $c$ and $z$ are both 16-dimensional. All FC layers are 64-dimensional and followed by a rectified linear unit (ReLU). The network weights for computing the spatial features, geometric features, and relation features are shared across modules.

\subsection{Baselines}
\begin{figure}[t!]
    \centering
    \includegraphics[width=\linewidth]{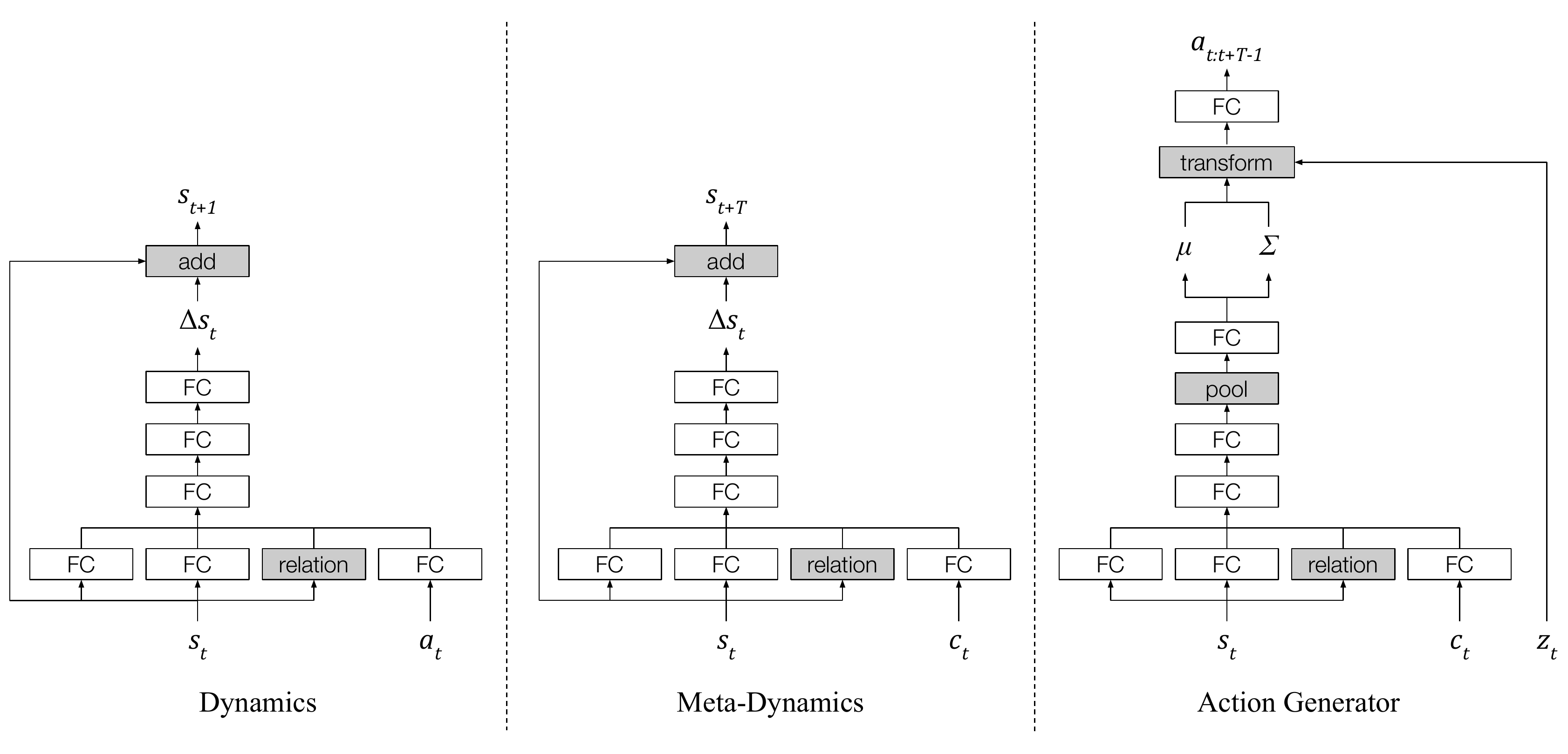}
    \caption{Modules of CAVIN Planner.}
    \label{fig:cavin_modules}
\end{figure}

\begin{figure}[t!]
    \centering
    \includegraphics[width=0.616\linewidth]{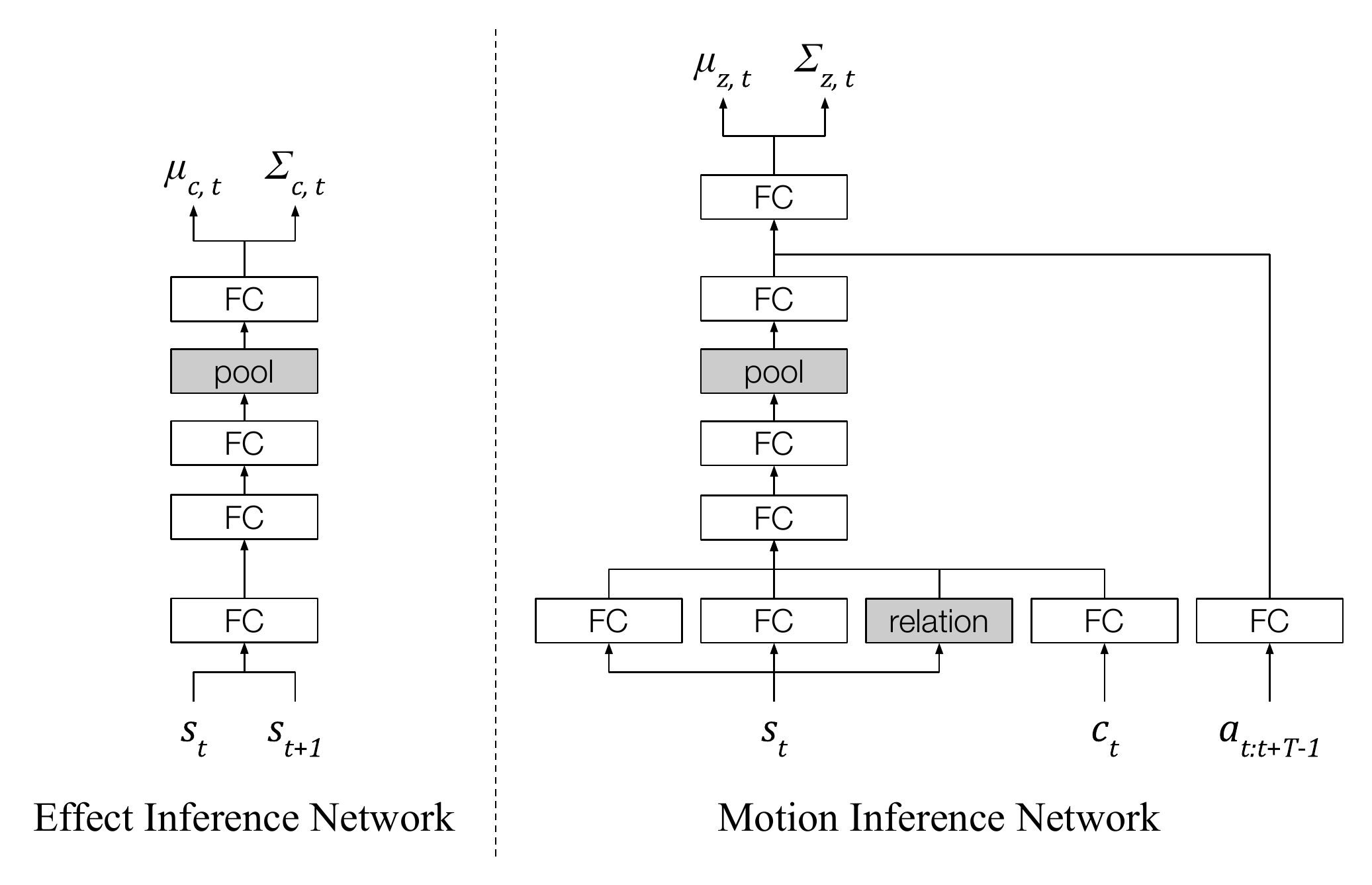}
    \caption{Inference Networks of CAVIN Planner.}
    \label{fig:inference_networks}
\end{figure}

\begin{figure}[t!]
    \centering
    \includegraphics[width=\linewidth]{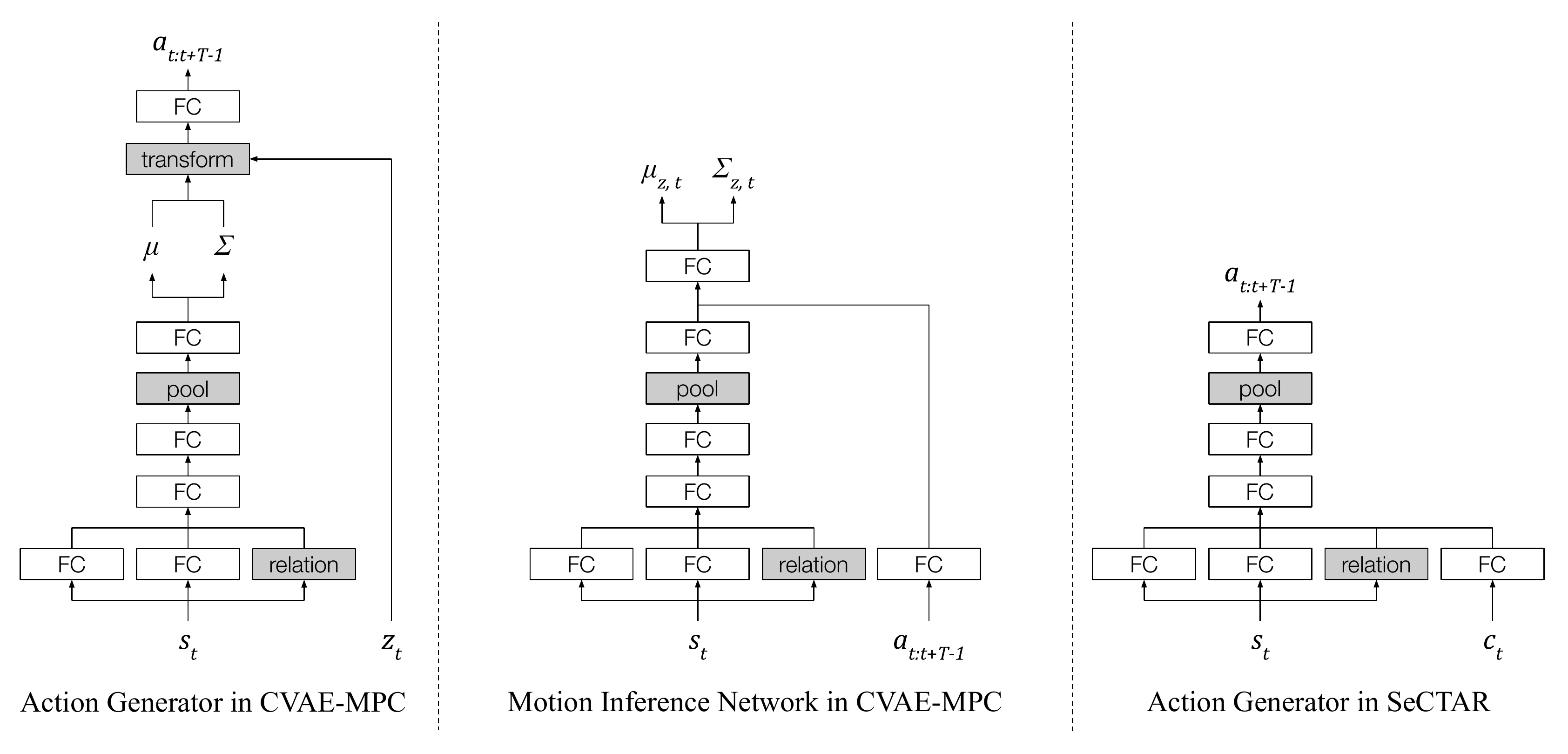}
    \caption{Modules and Inference Networks in Baselines.}
    \label{fig:baseline_adaptations}
\end{figure}

We adapt baseline methods to the multi-step manipulation tasks evaluated in this paper. Here we describe these baselines and compare with our model design in the context of this paper. \textbf{MPC}~\cite{guo2014deep, Agrawal2016LearningTP, Finn2017DeepVF} runs sampled-based planning by directly drawing samples from the original action space without a learned action generator. We adapt \cite{ichter2018learning} as \textbf{CVAE-MPC} to learn a conditional VAE to sample collision-free trajectories for motion planning without a high-level dynamics model. The CVAE is used to sample action sequences as our action generator. \textbf{SeCTAR}~\cite{co2018self} uses a VAE for jointly sampling trajectories states and actions from the latent space in a self-consistent manner. In SeCTAR, both state and actions are decoded from a single latent variable (which can be considered as the effect code $c$ in our case). In this way, the generative process of the states is equivalent to our high-level dynamics model, while the generative process of actions can be considered as a action generator given only $c$ but not $z$. To enable replanning every $T$ steps, we use the same action generator as in this paper to generate action sequences of $T$ steps at once instead recursively predicting for each time step as in \cite{co2018self}. Baselines use the same network architectures as in CAVIN for their counterparts except for modules in Fig.~\ref{fig:baseline_adaptations}. These different network designs are due to different module inputs in baselines. To this end, we design the adapted architectures by removing unused layers from Fig.~\ref{fig:cavin_modules} and Fig.~\ref{fig:inference_networks}. We use 32-dimensional latent codes for CVAE-MPC and SeCTAR to have fair comparisons with the CAVIN Plannar in which the latent codes are totally 32-dimensional. 

\subsection{Training}
To train all methods, we collect $500,000$ random transitions of planar pushing using a heuristic pushing policy. The dataset is partitioned for training (90\%) and validation (10\%) for hyperparameter tuning. Training terminates after one million steps when no further decrease in the total loss is observed on the validation set. We use L2 losses for reconstructing actions and states in the ELBO of cascaded variational inference. For states, we sum up the L2 losses of the positions and geometric features of all objects. We use Adam optimizer~\cite{kingma2014adam} with a learning rate of 0.0001.

\vspace{20pt}


\begin{figure}[t!]
    \centering
    \includegraphics[width=\linewidth]{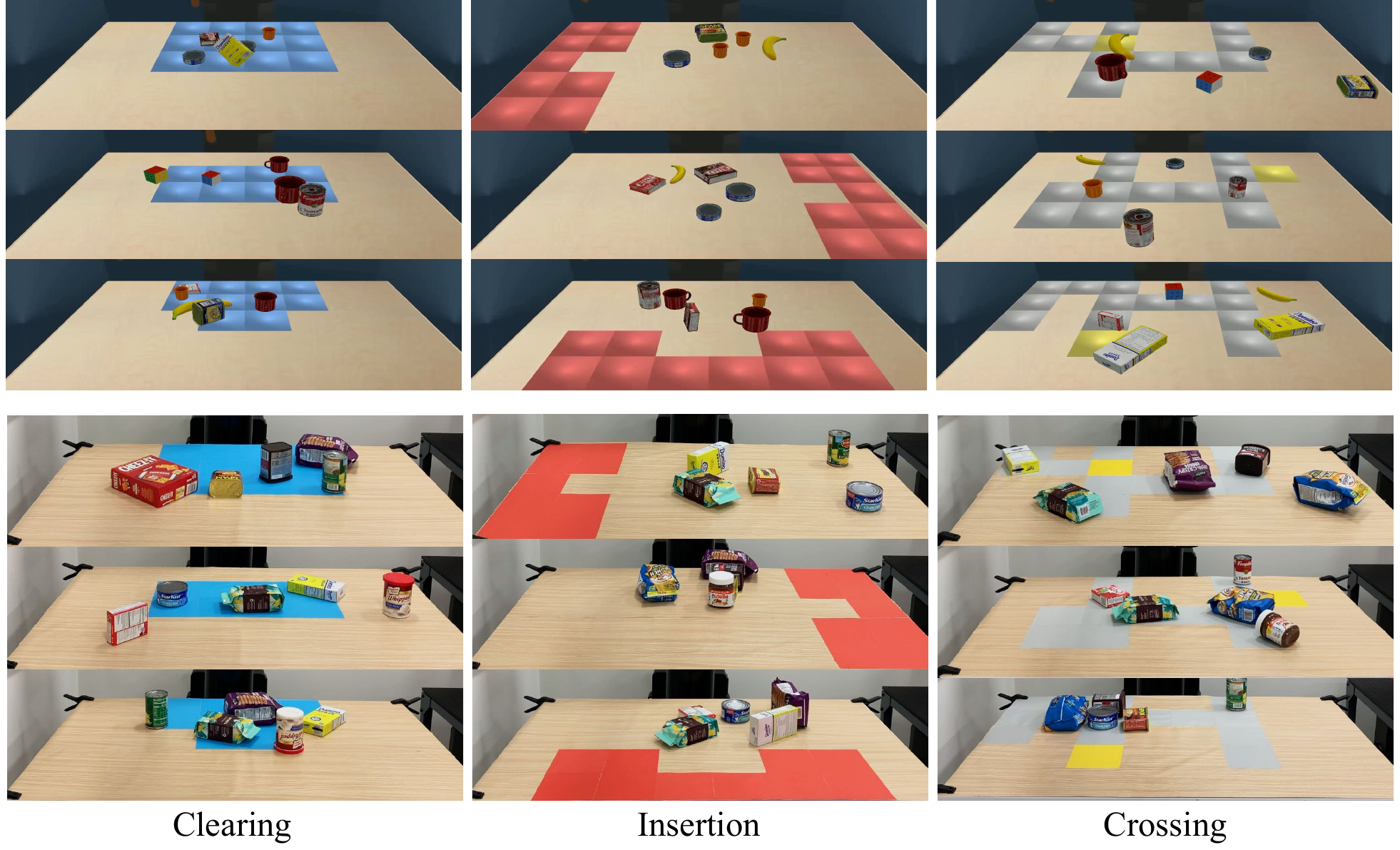}
    \caption{Task variations in simulation (top) and the real world (bottom).}
    \label{fig:task_layouts}
\end{figure}

\section{Task Details}

\begin{wrapfigure}{R}{0.4\textwidth}
\centering
\includegraphics[width=0.35\textwidth]{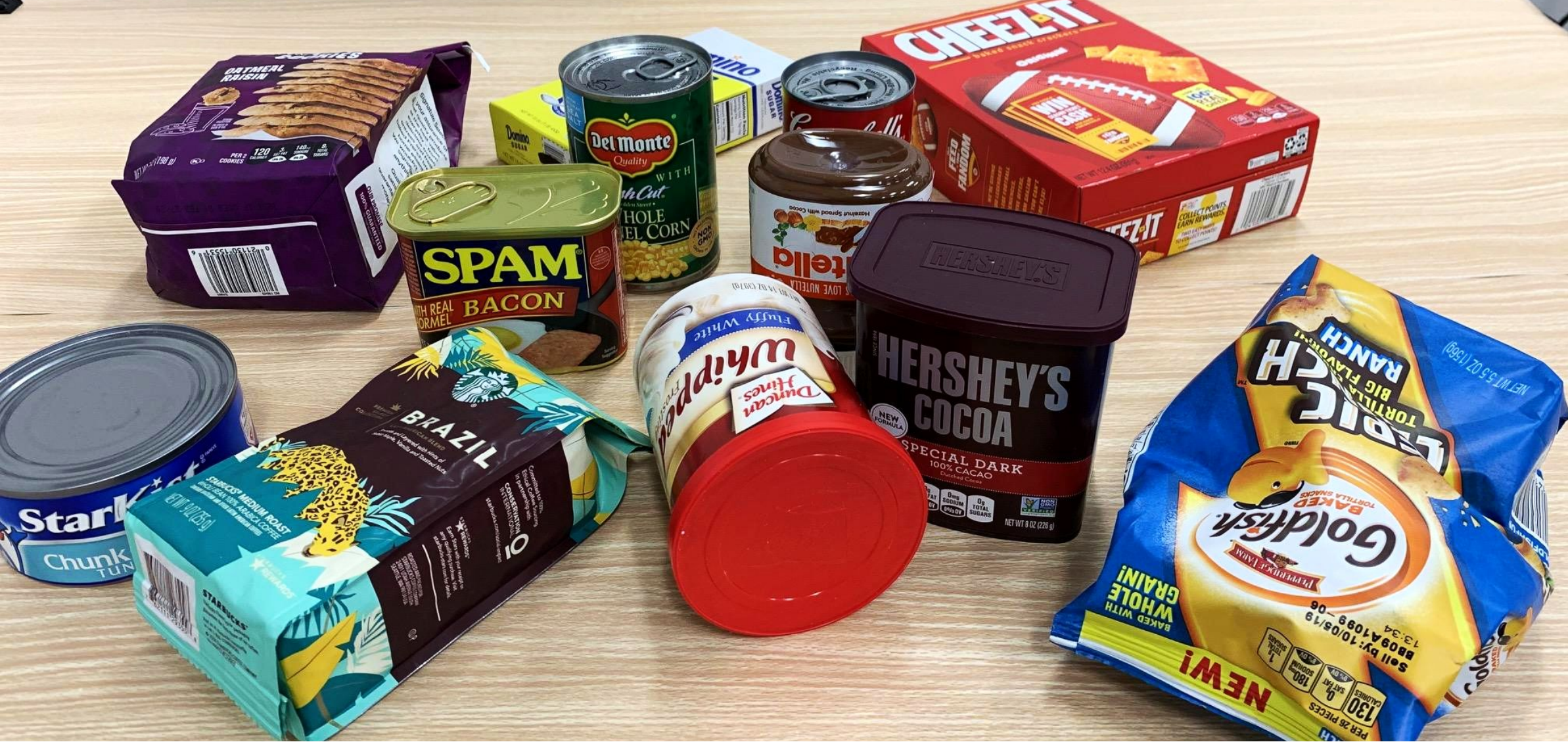}
\caption{Real-world object set.}
\label{fig:real_objects}
\end{wrapfigure}

In this section, we describe the details of the three multi-step manipulation tasks evaluated in the paper. The simulated and real-world environments for the three tasks are illustrated in Fig.~\ref{fig:tasks}. In Fig.~\ref{fig:task_layouts}, we show variations of objects and maps in simulation and the real world. We use all maps for qualitative analysis but only the maps on the top row for quantitative evaluation. In simulation, we select a subset of objects from YCB Dataset~\cite{alli2015TheYO} including \texttt{sugar\_box}, \texttt{tomato\_soup\_can}, \texttt{tuna\_fish\_can}, \texttt{pudding\_box}, \texttt{gelatin\_box}, \texttt{potted\_meat\_can}, \texttt{banana}, \texttt{mug}, \texttt{a\_cups}, \texttt{rubiks\_cube}. In the real world, we use the object set as shown in Fig.~\ref{fig:real_objects}.


\texttt{Clearing}: The goal is to clear all objects within a region denoted by blue masks on the table. The success is achieved when none of the objects remains in the region. 

\texttt{Insertion}: A slot is placed randomly along one side of the table, surrounded by restricted area denoted by red masks. A target object is specified while other movable objects are treated as distracting obstacles. The success is achieved when the target object is moved into the slot. The task fails if any object moves into the restricted area.

\texttt{Crossing}: A bridge denoted by grey tiles is placed on the table. A target object is randomly placed on a starting position on the bridge. The robot needs to move the target object to a goal position (in golden masks) on the other side of the bridge and moves away obstacles on the way. If the target object leaves the bridge the task fails. 

The position of an object is defined by the center of its point cloud. In addition to the task constraints mentioned above, we also terminates an episode when objects leave the workspace of the robot. The model receives a reward of 100 when reaching the goal in each of the task and a penalty of -100 when the episode terminates because of violation of the task constraints. A time penalty of -1 is added to the reward every step to encourage efficient plans. For dense rewards, we include the distance that the target object moves towards the goal position in terms of meters. Since there is no goal position in the clearing task, we set virtual goals at the edge of the tables to encourage the robot to move objects out of the masked region.

\end{document}


\maketitle

\renewcommand\thesection{\Alph{section}}



\section{Task Descriptions}
We use the following three multi-stage tabletop manipulation tasks to evaluate our proposed \algoName model against baselines. These tasks involve multi-object interactions with complex dynamics, which requires the robot to perform long-horizon reasoning to come up with an effective sequence of actions to complete each task from visual observations.

\texttt{Clearing}: The goal is to clear a work space region (in blue masks) on the table by moving away all objects within. Success is achieved when none of the objects remains in the workspace. 

\texttt{Insertion}: A slot is placed randomly along one of the sides of the table, surrounded by restricted area indicated in red masks. A target object is specified while other movable objects are treated as distracting obstacles. The success is achieved when the target object is moved into the slot. If any object moves into the restricted area the task fails. 

\texttt{Crossing}: A ``bridge'' is randomly constructed (in grey masks) on the table and the target object is placed on a random starting position on the bridge. The robot needs to move the target object to a goal position (in golden masks) on the other side of the bridge and moves away obstacles on the way. If the target object leaves the bridge the task fails.

\section{Baselines}
Here we describe each baseline and compare with our model design in the context of this paper.

\textbf{Visual MPC}~[9] runs sampled-based planning by directly drawing samples from the original action space without a learned action sampler. 

\textbf{CVAE-SBMP} adapts prior work [21] to learn a conditional VAE to sample collision-free trajectories for motion planning without a high-level dynamics model. We use their model to sample sequences of actions, same with our action sampler, to compare with our model.

\textbf{SeCTAR}~[24] uses a VAE for jointly sampling trajectories states and actions from the latent space in a self-consistent manner. In SeCTAR, both state and actions are decoded from a single latent variable (which can be considered as the effect code $c$ in our case). In this way, the generative process of the states is equivalent to our high-level dynamics model, while the generative process of actions can be considered as a action sampler given only $c$ but not $z$. Lastly, we also compare a variant of our \algoName trained without the counterfactual regularization (CFR) in Sec.~\ref{sec:training}. All model variants and baselines use the same network architecture for their module counterparts. We use 64-dimensional latent codes for CVAE-SBMP and SeCTAR and 32-dimensional $c$ and $z$ (totally 64-dimensional) to have comparisons using latent features of equivalent dimensions.


\begin{figure}[!h]
    \centering
    \includegraphics[width=1.0\linewidth]{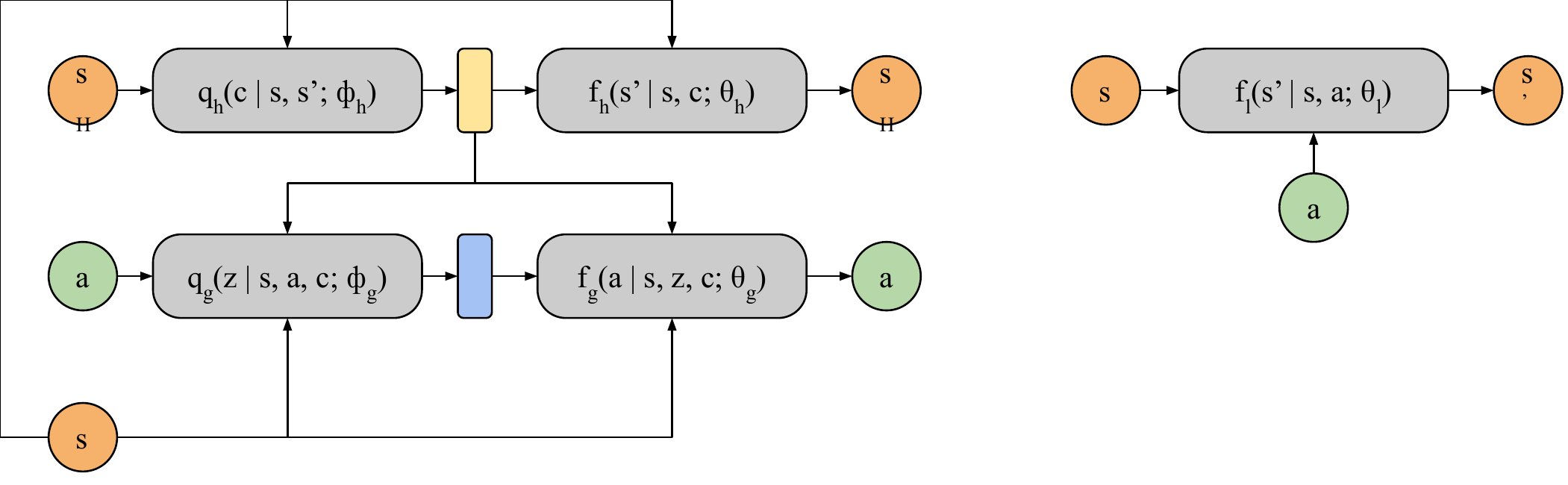}
    \caption{Model-Figure}
    \label{fig:diagram}
\end{figure}

\section{Model Diagram}
We include a computational diagram of the training procesess

\renewcommand*{\bibfont}{\footnotesize}